\documentclass[lettersize,journal]{IEEEtran}
\usepackage{amsmath,amsfonts}
\usepackage{algorithmic}
\usepackage[linesnumbered,ruled,vlined]{algorithm2e} 
\usepackage{array}
\usepackage[caption=false,font=normalsize,labelfont=sf,textfont=sf]{subfig}
\usepackage{textcomp}
\usepackage{stfloats}
\usepackage{url}
\usepackage{multicol}
\usepackage{multirow}
\usepackage{verbatim}
\usepackage{graphicx}
\usepackage{cite}

\begin{document}

\title{Improved Obstacle Avoidance for Autonomous Robots with ORCA-FLC\\
}

\author{\IEEEauthorblockN{Justin London \\}
\IEEEauthorblockA{\textit{Dept. of Electrical Engineering and Computer Science} \\
\textit{University of North Dakota}\\
Grand Forks, North Dakota \\
justin.london@und.edu}}

\maketitle

\begin{abstract}
    Obstacle avoidance enables autonomous agents and robots to operate safely and efficiently in dynamic and complex environments, reducing the risk of collisions and damage. For a robot or autonomous system to successfully navigate through obstacles, it must be able to detect such obstacles.  While numerous collision avoidance algorithms like the dynamic window approach (DWA), timed elastic bands (TEB), and reciprocal velocity obstacles (RVO) have been proposed, they may lead to suboptimal paths due to fixed weights, be computationally expensive, or have limited adaptability to dynamic obstacles in multi-agent environments.  Optimal reciprocal collision avoidance (ORCA), which improves on RVO, provides smoother trajectories and stronger collision avoidance guarantees. We propose ORCA-FL to improve on ORCA by using fuzzy logic controllers (FLCs) to better handle uncertainty and imprecision for obstacle avoidance in path planning.  Numerous multi-agent experiments are conducted and it is shown that ORCA-FL can outperform ORCA in reducing the number of collision if the agent has a velocity that exceeds a certain threshold.  In addition, a proposed algorithm for improving ORCA-FL using fuzzy Q reinforcement learning (FQL) is detailed for optimizing and tuning FLCs.
\end{abstract}


\section{Introduction}
\ \ \ Obstacle avoidance is essential in path planning and navigation tasks.  Various obstacle avoidance algorithms exist such as dynamic window algorithms (DWA) and velocity field histograms (VFH).  Reciprocal velocity obstacles (RVO) is another algorithm that is designed to prevent multiple agents from colliding with each other while moving towards their goals.   RVO extends the concept of velocity obstacles (VO) \cite{Fiorini:1998} by defining the set of velocities that, if chosen by both agents, will ensure either agent has to significantly change its course to avoid the other.
RVO was introduced to overcome the oscillation motion in velocity obstacles (VO) where two agents cycle back and forth between their current velocity and old velocities to avoid each other \cite{Berg:2008}

RVO has a primary advantage over other these other algorithms in that it allows for decentralized planning: each agent makes decisions only based on the state of its neighbors without the need for a central controller.  Consider two agents A and B moving in a 2D or 3D plane.  Each agent seeks to reach its destination without colliding with other.   The Velocity Obstacle (VO) for A respect to B is the set of all relative velocities of A with respect to B that would result in a collision at some point in the future assuming agent B keeps its current velocity.  

RVO extends the concept by defining the set of velocities that, if chosen by both agents, will ensure either agent has to significantly change its course to avoid the other. RVO assumes that both Robot A and B make an equal effort to avoid each other based on their set of relative velocities that will lead to a collision if both robots continue on their path.

One of the limitations of RVO path planning is its potential to generate oscillations, especially in complex environments with many agents and dynamic obstacles.  The RVO assumes that all agents react and respond similarly to their surroundings, which is not always accurate, leading to unpredictable and sometimes undesirable movement patterns given different agents may have different goals, decision-making processes, or priorities.  RVO assumes perfect knowledge of the environment, including the position, radius, and velocity of other agents and obstacles.   This is often not realistic in real-world environments where information is limited or incomplete.    
 
RVO assumes that both Robot A and B make an equal effort to avoid each other based on their set of relative velocities that will lead to a collision if both robots continue on their path.  Thus, RVO implicitly assumes that each agent takes equal responsibility in the effort to avoid mutual collisions, but can be generalized to allocate a different share balance between agents by allowing agent A to avoid agent B by using a share $\alpha$ such that their velocity is a weighted average of their velocity and 1 - $\alpha$ of the other agent's velocity.  RVO works well in one-on-one obstacle environments, but not with multiple agents.

van den Berg, et. al. \cite{Berg:2008} introduced optimal reciprocal collision avoidance (ORCA), building on VO and RVO model.  ORCA computes and optimal collison-free velocity plane (in the velocity space) for each moving obstacle.  The robot then selects its optimal velocity from the intersection of all collision-free planes.  ORCA uses the relative velocity of agent to determiine their admissible velocity domains.  The radius of A and B are added so that agent A must never be inside the circle or disk $D$(B,R) centered at B with radius R.

\section{Related Works}

    Other dynamic obstacle avoidance path planning algorithms include the dynamic window approach, timed elastic band, and adaptive Monte Carlo localization.   The Timed Elastic Band (TEB) algorithm was introduced to resolve collision avoidance in real-time during robot motion planning \cite{Rosmann:2012, Quinlan:1993}  The TEB algorithm solves the optimal sequence of robot poses with timed intervals.   
    The elastic band method optimizes the robot trajectory by modifying the initial trajectory generated by the global path planner in real-time to generate a collision-free path while maintaining the closest global point and continuously deforming it to maintain distance from the obstacle.  The TEB algorithm enhances the elastic band method by converting the initial path, consisting of a series of path waypoints, into a time-dependent trajectory, explicitly considering temporal information about the motion such as dynamic constraints on robot speed and acceleration.     
    
    Wang et. al. \cite{Wang:2023} proposed TEB-CA, incorporating the obstacle avoidance velocity $v^{ORCA}_{A}$ predicted by the ORCA model into the total cost function of the TEB algorithm to obtain the optimal collision avoidance trajectory.  They apply their approach to a four-wheeled car-like autonomous mobile robot with Ackermann steering kinematics.  Chauvin, Gianazza, and Durand \cite{Durand:2024} combine ORCA with A* path-planning algorithm as a reactive method for drones or unmanned aerial systems (UAVs) flying in dense urban environments.   Chen et. al. \cite{Cheng:2017} used ORCA with Model Predictive Control (MPC).  The ORCA algorithm enables each agent to determine its permitted velocity and the MPC method anticipates future trajectories of a system based on its system dynamics and numerous constraints. 

    ORCA hybrids have also been extended to multiple-agent environments.  Roncero, Muchacho, and Ogren \cite{Roncero:2024} proposes a combination of Velocity Obstacles (VO) and Control Barrier Functions (CBF) for multi-agent collision avoidance. Hybrid obstacle avoidance algorithms (HVRO) have been proposed. Giese \cite{Giese:2014} introduced the notion of rotation with Reciprocally-Rotating Velocity Obstacles (RRVO) extending ORCA for local collision avoidance to help mitigate potential deadlock scenarios that arise when using polygonal agents.   

    Romlay, Ibrahim, et. al. \cite{Romlay:2023} introduced Fuzzy Logic Controller-ORCA (FLC-ORCA) to enhance obstacle avoidance for robotic navigation aids (RNAs) that assist visually impaired individuals.  They applied fuzzy logic controllers to determine the shared responsibility and predicted velocity parameters.  We build upon and extend their work by making the following contributions: (1) incorporating sensor fusion (2) fine tuning the FLCs; and (3) introducing reinforcement learning into the model as feedback into the optimization, and proposing ORCA-FL-RL.   
    
    In Section III, we provide a mathematical formulation of the model that serves as the foundation for the methodology used to implement the algorithm in numerous experiments and simulations. In section IV, the the methodology is detailed including the development of the FLCs and fuzzy logic rules. In section VI, we provide the experimental and simulation results.  In section VII, FLC optimization and tuning using fuzzy reinforcement Q-learning is detailed. In Section VIII, we conclude and provide a direction for future research in section IX.

\section{Mathematical Formulation}

\begin{figure}[h]
    \centering
	\begin{center} 
	\includegraphics[width=0.9\linewidth]{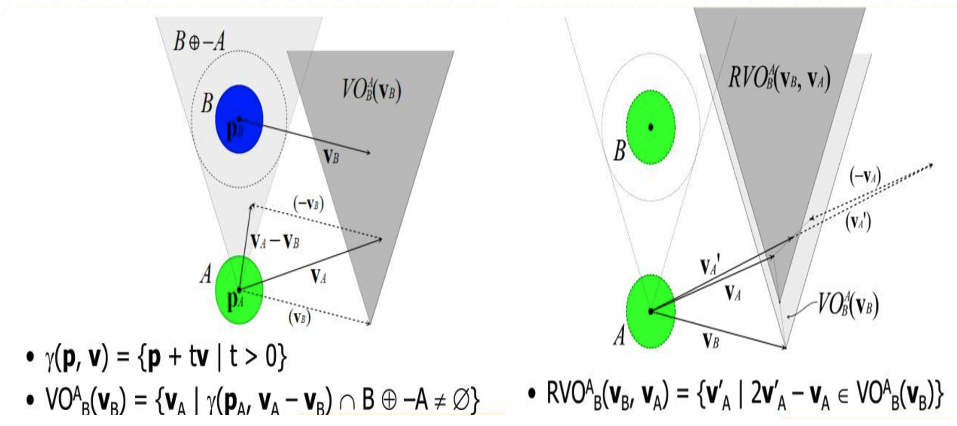}
	\caption{Left: Velocity Obstacle $VO^{A}_{B}(\boldsymbol{v}_{B})$ of a disc-shaped obstacle B to a disc-shaped agent A. Right:  The Reciprocal Velocity Obstacle $RVO^{A}_{B}(\boldsymbol{v}_{B},\boldsymbol{v}_{A})$ of agent B to agent A
    \cite{Berg:2008}.} 
	\label{fig:rvo} 
	\end{center}  
\end{figure}

    Let $\boldsymbol{v}_{A}$ and $\boldsymbol{v}_{B}$ be the velocities of agents A and B, respectively..  The relative velocity of A with respect to B is given by:
    \begin{equation}
        \boldsymbol{v}_{rel} = \boldsymbol{v}_{A} - \boldsymbol{v}_{B} 
    \end{equation}
    Let $\boldsymbol{p}_{AB} = \boldsymbol{p}_{A} - \boldsymbol{p}_{B}$ be the current relative position of A with respect to B and let $r_{AB} = r_{A} + r_{B}$ be the sum of their radii.  
    The VO for A with respect to B is the set of all relative velocities that will lead to a collision between A and B within a specified time horizon $\tau$.  Mathematically, it is defined as:
    \begin{multline}
        VO_{AB}(\tau) = \{\boldsymbol{v} \lvert \ \exists \ t \in [0,\tau]: \\ \frac{\boldsymbol{v}t - \boldsymbol{p}_{B} + \boldsymbol{p}_{A}}{t} \leq r_{A} + r_{B} \}
    \label{eq:eq1}
    \end{multline}.  
    \normalfont
    Equation \ref{eq:eq1} can be rewritten as 
    \begin{equation}
        \big \lvert \big \rvert \boldsymbol{p}_{AB} + \boldsymbol{v}_{AB}t \big \lvert \big \rvert < r_{AB}
    \end{equation}
    Dividing both sides by $t$, are rearranging yields:
    \begin{equation}
        \big \lvert \big \rvert \boldsymbol{v}_{AB} - (-\frac{\boldsymbol{p}_{AB}}{t}) \big \lvert \big \rvert < \frac{r_{AB}}{t}
    \end{equation}
    Thus, the velocity obstacle $VO_{AB}$ is the union of discs:
    \begin{equation}
        VO_{AB} = \underset{0 < t \leq \tau}{\bigcup} \ D(-\frac{\boldsymbol{p}_{AB}}{t},\frac{r_{AB}}{t} \big )
    \end{equation}
    Alternatively, the VO can be defined as:
    \begin{equation}
        VO_{AB}(\boldsymbol{v}_{B}) = \{ \boldsymbol{v}_{A} \ \lvert \lambda(\boldsymbol{p}_{A},\boldsymbol{v}_{A} -\boldsymbol{v}_{B}) \cap  B \oplus -A \neq 0 \}
    \end{equation}
    where $\lambda(\boldsymbol{p},\boldsymbol{v})$ is the ray starting at $\boldsymbol{p}$ and heading in the direction of the relative velocity $\boldsymbol{v}_{rel}$ of A and B (or $\boldsymbol{v}_{A} - \boldsymbol{v}_{B}$), intersects the Minkowski sum ($\oplus$) of B and -A centered at the $\boldsymbol{p}_{B}$ velocity $\boldsymbol{v}_{A}$ is in the velocity obstacle B.
    \begin{equation}
        \lambda(\boldsymbol{p},\boldsymbol{v}) = \{\boldsymbol{p} + t\boldsymbol{v} \ \lvert \ t \geq 0 \} 
    \end{equation}
    The RVO is defined as the Minkowski sum of the VO and the velocity of Robot B, shifted by half of the relative velocity:
    \begin{equation}
        RVO_{AB}(\tau) = VO_{AB}(\tau) + \frac{1}{2}\boldsymbol{v}_{rel}
    \end{equation}
    Given two robots A and B in space (2D or 3D) each robot has a current position P, a goal position G, a defined speed $s$, and a defined radius $r$.   The objective is to move each robot towards its goal while avoiding collisions with other robots.   
    The relative position for any two robots A and B, the relative position is given by:
    \begin{equation}
            \boldsymbol{R}_{AB} = \boldsymbol{P}_{B} - \boldsymbol{P}_{A}
    \end{equation}
    The Euclidean distance DAB between robot A and B is:
\begin{equation} 
    D_{AB} = \lvert \rvert \boldsymbol{R}_{AB} \lvert \rvert = \sqrt{R^2_{AB_x} + R^2_{AB_y} + R^2_{AB_{z}}}
\end{equation}
    If the distance between A and B, $D_{AB}$ is less than the sum of the radii of the two robots and their speeds
    \begin{equation}
        D_{AB} < r_{A} + r_{B} + s
    \end{equation}
    then an avoidance vector $\boldsymbol{A}$ is computed using the cross product:
    \begin{equation}
        \boldsymbol{A} = \boldsymbol{R}_{AB} \times \boldsymbol{U}
    \end{equation}
    In essence, velocity obstacle avoidance methods rely on the intersection of velocity obstacles, also known as collision cones (CC), from each agent emanating from their current position P.   Figure \ref{fig:rvo} illustrates the geometry of VOs and RVOs. 

    To overcome oscillations and to make VO more robust in dynamic environments, hybrid RVO was proposed \cite{Snape:2009}.

    In ORCA, both agents are assumed to equally share the collision avoidance responsibility, reciprocally collision-free for at least $\tau$ time, have maximal permitted velocities, and do not communicate, but can infer one another's trajectory path.  Denote the sets of velocities that obey these assumptions for A as $ORCA^{\tau}_{A|B}$ and B as $ORCA^{\tau}_{B|A}$.  Denote the pair of velocities that maximize the permitted velocities that are close to "optimization" velocities as $\boldsymbol{v}^{opt}_{A}$ and $\boldsymbol{v}^{opt}_{B}$ and assume that  $\boldsymbol{v}^{opt}_{A} - \boldsymbol{v}^{opt}_{B} \in VO^{\tau}_{A|B}$.    

\begin{figure}[h]
    \centering
	\begin{center} 
	\includegraphics[width=0.9\linewidth]{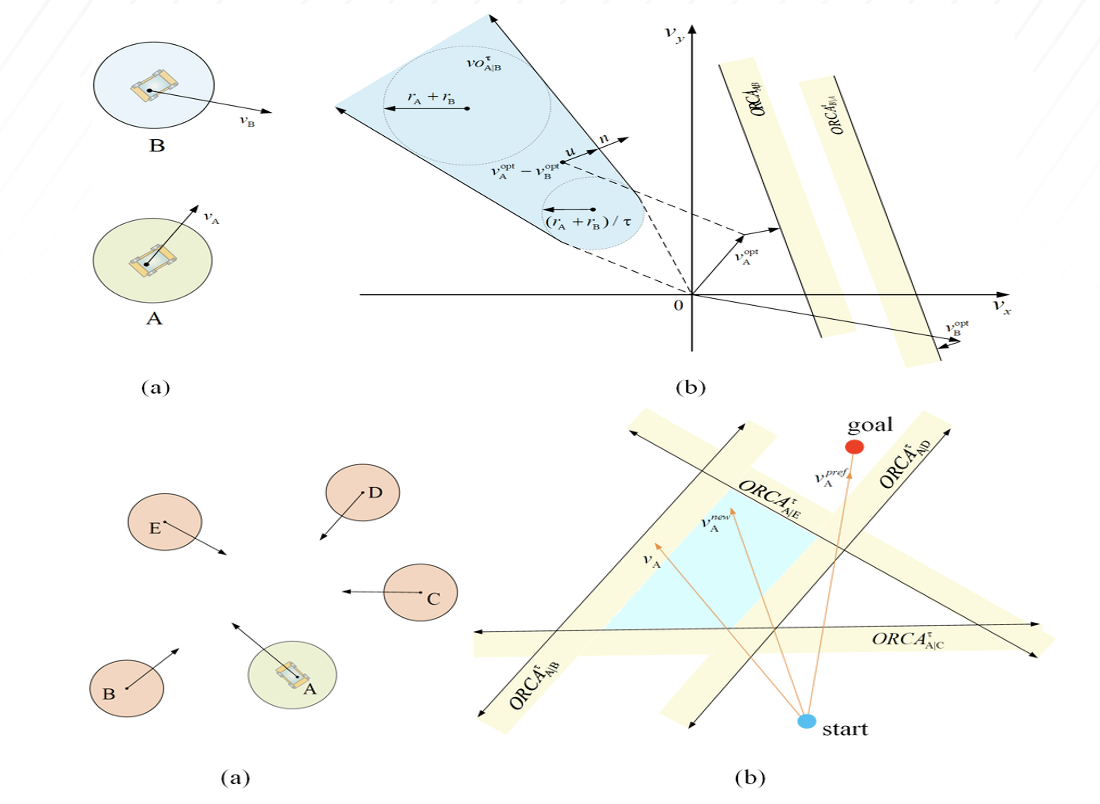}
	\caption{Top left (a): A workspace configuration with two agents. Top right (b): $VO^{\tau}_{A|B}$ and the set of $ORCA^{\tau}_{A|B}$ (half-plane) of permitted velocities for A.  For holonomoic robots, $\xi$ = 0.  Bottom left (a):  The ORCA lines for robots navigation in dynamic environments.  Bottom right (b): The available choices of collision-free velocity space are shown \cite{Zhang:2015}.} 
	\label{fig:orca} 
	\end{center}  
\end{figure}

    Let $\textbf{u}$ be the vector from $\textbf{v}^{opt}_{rel} = \textbf{v}^{opt}_{A} - \textbf{v}^{opt}_{B}$  to the nearest point on the velocity obstacle boundary such that
    \begin{equation}
        \textbf{u} = (\underset{\textbf{v} \in \partial VO^{\tau}_{A|B}} {\text{arg min}}(\lvert \rvert \textbf{v}-\textbf{v}^{opt}_{rel} \lvert \rvert) - \textbf{v}^{opt}_{rel}
    \end{equation}
    and let $\textbf{n}$ be the outward normal of the boundary of the velocity obstacle, $\partial VO^{\tau}_{A|B}$ at point $\textbf{v}^{opt}_{rel} + \textbf{u}$ where $\textbf{u}$ is the smallest change required to the relative velocity of A and B to avert collision within $\tau$ time.  To share the responsibility of avoiding collisions, robot A adapts its velocity by (at least) $\frac{1}{2}\textbf{u}$ and assumes that B takes care of the other half. Therefore, the set $ORCA^{\tau}_{A|B}$ of permitted velocities for A is the half-plane pointing in the direction of $\textbf{n}$ starting at point $\textbf{v}^{opt}_{rel} + \frac{1}{2}\textbf{u}$.  Then 
    \begin{equation}
        ORCA^{\tau}_{A|B} = \{\textbf{v} \lvert (\textbf{v} - (v^{opt}_{A} + \frac{1}{2}\textbf{u}))\cdot \textbf{n} \geq 0 \}
    \end{equation}
    The set $ORCA^{\tau}_{B|A}$ is defined symmetrically. However, in practice it not realistic to assume that each robot shares equal responsibility and therefore the responsibility allocation is dynamic at each time step based on the robots' relative velocities and positions.  A fuzzy logic controller can be used to determine what the avoidance responsibility should be based on the obstacle's velocity, acceleration, and distance from the robot. 
    \begin{equation}
        ORCA^{\tau}_{A|B} = \{\textbf{v} \lvert (\textbf{v} - (\textbf{v}^{opt}_{A} + \boldsymbol{x}\boldsymbol{u}))\cdot \textbf{n} \geq 0)\}
        \label{eq:orca}
    \end{equation}
    Figure \ref{fig:orca} shows a graphical illustration of how in a multi-agent environment, each robot generates an ORCA.  The area created by the intersecting ORCAs for each agent is an admissible collision-free velocity space. 

\begin{algorithm}
	\label{alg:rrtstaralg}
	\scriptsize
	\algsetup{linenosize=\small}
	\caption{\small \textbf{ORCA-FL}} 
    $\textbf{Input} \ \mathcal{A} : \text{List of robots}, \mathcal{O}$ : 
    $\text{List of obstacles}, \text{fuzzy logic controllers}: FLC1, FLC2$ \\
    $\text{Set} \ \Delta t$ \\
    $s \gets 0.5$ \\       
    \For{$\mathcal{A}_{i} \in \mathcal{A}$} {  
        $\text{Sense} \ \boldsymbol{p}_{i} \ \text{and} \ \boldsymbol{v}_{i}$ \\
        $\text{Get} \ \text{nearest neighbor} \ A_{j}$ \\
	    $\textbf{x}_{i} \gets {Pos}_{A_i}$ \\
        $\textbf{x}_{j} \gets {Pos}_{A_j}$ \\
        $\textbf{v}_{i} \gets {Vel}_{A_i}$ \\
        $\textbf{v}_{j} \gets {Vel}_{A_j}$ \\
        \For{$A_{j} \in \mathcal{A} \ i \neq j$} { 
            $\text{Construct} \ VO_{A_{i}\lvert A_{j}} = \{\boldsymbol{v} \lvert \ \exists \ t \in [0,\tau]: tv \in D(\boldsymbol{p}_{A_{j}} + \boldsymbol{p}_{A_{i}}, r_{A_{i}} + r_{A_{j}})\}$  \\
            $\text{Compute} \ \textbf{n} \ \text{outward normal to closest point to the boundary of} \ VO^{\tau}_{A_{i} \lvert A_{j}}$ \\ $\text{at point} (\textbf{v}^{opt}_{A_i} - \textbf{v}^{opt}_{A_j}) + \textbf{u}$ \\
            $\text{Decide shared factor of avoidance responsibility} \ \textbf{s} \ \text{from} \ FLC1$ \\
            $\text{Construct} \ ORCA^{\tau}_{A_{i} \lvert A_{j}} \gets (\boldsymbol{v}_{i} - (\boldsymbol{v}^{opt}_{A_{i}} + s\boldsymbol{u}))\cdot \boldsymbol{n}$ \\
            $\textbf{v}^{rel} \gets v_{i} - v_{j}, \ i \neq j$ \\
            $\text{Compute change to avert collision} \ \textbf{u} = \underset{\textbf{v} \in \partial VO^{\tau}_{A_{i}|A_{j}}}{\text{arg min}}(\lVert \textbf{v}- (\textbf{v}^{opt}_{rel} - \textbf{y}) \rVert ) - (\textbf{v}^{opt}_{rel} - \textbf{y})$ \\
            $\text{Decide predicted velocity} \ \textbf{y} \gets \ v^{pvel}_{A_{j}} \ \text{from} \ FLC2$ \\
            $\text{Compute new velocity for} \ A_{i} \in ORCA^{\tau}_{A} = D(0,v^{max}_{A})\cap (\underset{A_{i} \neq A_{j}}{\cap} \ ORCA^{\tau}_{A_{j}})$ \\  
            $\text{Compute new velocity} \ v^{new}_{A_i} = \underset{\textbf{v}_{i} \in ORCA^{\tau}_{A_{j}}}{\text{arg min}}{\lvert \rvert \textbf{v} - \textbf{v}^{new}_{A_{i}} \lvert \rvert}$ \\
            $\text{Compute new position} \ p^{new}_{A_{i}} = p_{A_{i}} + v^{new}_{A_{i}}\Delta t$ \\
        }
        $\textbf{end}$ \\
    }
    $\textbf{end}$ 
\end{algorithm}

\section{Methodology}
    A fuzzy inference system (FIS) is a system that uses fuzzy set theory to map inputs to outputs. Two fuzzy logic controllers are implemented using the Fuzzy Logic Design Toolkit in Matlab.  The velocity, distance, and acceleration of the nearest obstacle to a given agent are fuzzified and input into Fuzzy Logic Controller 1 (FC1).  The FLC1 membership function (MF) inputs are distance are very near (VN), near (N), far (F), very far (VF).  The MF inputs for velocity are very slow (VS), slow (S), fast (F), very fast (VF).  The MF inputs for acceleration are deaccelerate (DCC), zero, and accelerate (ACC).  FLC1 has 1 output, and 48 rules.  The FLC1 output is the share of avoidance responsibility \textbf{u} (in \ref{eq:orca} with values that vary from 0 to 1 ($a = 0, b = 0.14, c = 0.29, d=0.43, e=0.57, f=0.71, g=0.86, h=1$) 

    FLC2 outputs the predicted or expected velocity of the nearest agent/obstacle in the next scanning cycle using the obstacle's velocity, density of the surroundings (i.e., number of obstacles the robot senses and detects in the vicinity), and the obstacle's acceleration as inputs.  The FLC2 inputs for MF velocity are very slow (VS), slow (S), fast (F), and very fast (VF).  The MF inputs for density are very low (L), medium (M), and high (H).  The MF inputs for the obstacle's acceleration are low, medium, and high.   FLC2 has 3 inputs, 1 output, and 27 rules.   The Mandami Type I implication/inference operator is used to convert the linguistic rules that describe how the FIS should make a decision regarding classifying an input or controlling an output.    Table \ref{tbl:table1} and Table \ref{tbl:table2} show the inputs, output, and membership functions for FLC1 and FLC2, respectively.  \textit{Trimf} denotes the triangle membership function.

\begin{figure}[h]
    \centering
	\begin{center} 
	\includegraphics[width=0.9\linewidth]{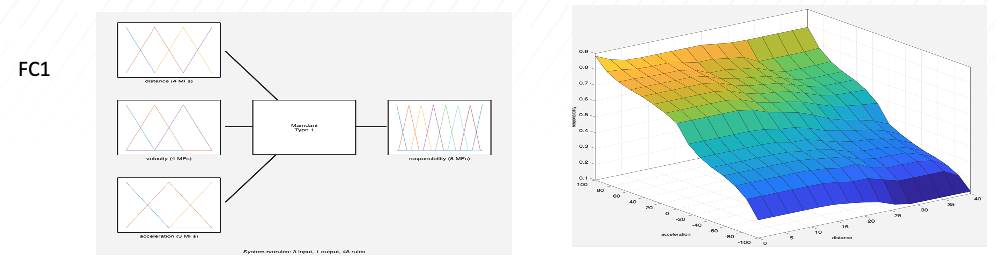}
	\caption{Fuzzy Logic Controller 1.  Left: Mandami Type I FIS.  Right: Control surface.} 
	\label{fig:flc1} 
	\end{center}  
\end{figure}

\begin{figure}[h]
    \centering
	\begin{center} 
	\includegraphics[width=0.9\linewidth]{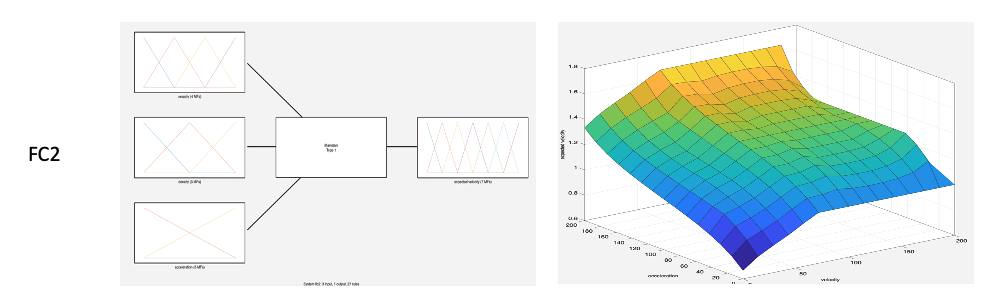}
	\caption{Fuzzy Logic Controller 2.  Left: Mandami Type I FIS.  Right: Control surface.} 
	\label{fig:flc4} 
	\end{center}  
\end{figure}

    Figures \ref{fig:flc1} and \ref{fig:flc4} are graphical representations of FIS 1 and FIS 2, respectively.  The right hand side of each plot show the 3D control surface of the FIS.  The fuzzy logic rules are IF-THEN rules.  For instance, using the fuzzy rule base of FLC1, Rule 1 provides IF \textit{velocity} is VS and \textit{distance} is VF and \textit{acceleration} is DCC, THEN \textit{responsibility} is a.  Using the fuzzy rule base of FLC2, Rule 2 is IF \textit{velocity} is VF and \textit{density} is L and \textit{acceleration} is ACC, THEN the \textit{predicted velocity} is g.
\begin{table}
\centering
\scriptsize
\caption{FLC1 Membership Functions}
\label{tbl:table1}
\begin{tabular}{|l|l|l|l|}\hline
  \multirow{10}{*}{Input} 
  & \multirow{4}{*}{Distance} & & \\ 
  & & \multirow{1}{*}{Very Near} & Trimf[-33 0 33] \\
  & & \multirow{1}{*}{Near} & Trimf[0 33 66] \\  
  & & \multirow{1}{*}{Far} & Trimf[33 66 100] \\  
  & & \multirow{1}{*}{Very Far} & Trimf[66 100 100] \\ 
  & & &   \\ \cline{2-4}
  & \multirow{4}{*}{Velocity} &  & \\
  & & \multirow{1}{*}{Very Slow} & Trimf[-66.67 0 66.67] \\
  & & \multirow{1}{*}{Slow} & Trimf[0 66.67 133.33] \\  
  & & \multirow{1}{*}{Fast} & Trimf[66.67 133.33 200] \\  
  & & \multirow{1}{*}{Very Fast} & Trimf[66.67 133.33 200] \\ 
  & & &   \\ \cline{2-4}
  & \multirow{4}{*}{Acceleration} & & \\ 
  & & \multirow{1}{*}{Dcc} & Trimf[-500 -250 0] \\
  & & \multirow{1}{*}{Zero} & Trimf[-100 0 100] \\  
  & & \multirow{1}{*}{Acc} & Trimf[0 250 500] \\  
  & & &    \\ \cline{1-4}
  \multirow{10}{*}{Output} 
  & \multirow{4}{*}{Collision Avoidance} & & \\ 
  & & \multirow{1}{*}{a} & Trimf[-0.119 0 0.119] \\
  & & \multirow{1}{*}{b} & Trimf[0.0238 0.1429 0.2619] \\  
  & & \multirow{1}{*}{c} & Trimf[0.1667 0.2857 0.4048] \\  
  & & \multirow{1}{*}{d} & Trimf[0.3095 0.4286 0.5476] \\ 
  & & \multirow{1}{*}{e} & Trimf[0.4524 0.5714 0.6905] \\
  & & \multirow{1}{*}{f} & Trimf[0.5952  0.7142 0.8333] \\
  & & \multirow{1}{*}{g} & Trimf[0.7381 0.8571 0.9762] \\
  & & \multirow{1}{*}{h} & Trimf[0.81 1 1.1119] \\
  & & &   \\ \cline{1-4}
\end{tabular}
\end{table}

\begin{table}
\centering
\scriptsize
\caption{FLC2 Membership Functions}
\label{tbl:table2}
\begin{tabular}{|l|l|l|l|}\hline
  \multirow{10}{*}{Input} 
  & \multirow{4}{*}{Velocity} &  & \\
  & & \multirow{1}{*}{Very Slow} & Trimf[-66.67 0 66.67] \\
  & & \multirow{1}{*}{Slow} & Trimf[0 66.67 133.33] \\  
  & & \multirow{1}{*}{Fast} & Trimf[66.67 133.33 200] \\  
  & & \multirow{1}{*}{Very Fast} & Trimf[66.67 133.33 200] \\ 
  & & &   \\ \cline{2-4}
  & \multirow{4}{*}{Density} & & \\ 
  & & \multirow{1}{*}{Low} & Trimf[-4 0 4] \\
  & & \multirow{1}{*}{Medium} & Trimf[0 4 8] \\  
  & & \multirow{1}{*}{High} & Trimf[4 8 12] \\  
  & & &   \\ \cline{2-4}
  & \multirow{4}{*}{Acceleration} & & \\ 
  & & \multirow{1}{*}{Dcc} & Trimf[-500 -250 0] \\
  & & \multirow{1}{*}{Zero} & Trimf[-100 0 100] \\  
  & & \multirow{1}{*}{Acc} & Trimf[0 250 500] \\  
  & & &    \\ \cline{1-4}
  \multirow{10}{*}{Output} 
  & \multirow{4}{*}{Expected Velocity} & & \\ 
  & & \multirow{1}{*}{a} & Trimf[-0.333 0 0.333] \\
  & & \multirow{1}{*}{b} & Trimf[0 0.333 0.6667] \\  
  & & \multirow{1}{*}{c} & Trimf[0.3333 0.6667 1] \\  
  & & \multirow{1}{*}{d} & Trimf[0.6667 1 1.333] \\ 
  & & \multirow{1}{*}{e} & Trimf[1 1.333 1.6667] \\
  & & \multirow{1}{*}{f} & Trimf[1.333 1.6667 2] \\
  & & \multirow{1}{*}{g} & Trimf[1.6667 2 3.6667] \\
  & & &   \\ \cline{1-4}
\end{tabular}
\end{table}
    The outputs of FLC1 and FLC2 are converted to crisp values through defuzzification using the centroid method.  For each experiment, we track the number of iterations (time steps), total elapsed time, number of collisions, maximum speed, maximum acceleration, sensor range, and number of dynamic obstacles.  We compare the results of ORCA with ORCA-FL.


\section{Experiments}
\begin{figure}[h]
    \centering
	\begin{center} 
	\includegraphics[width=0.9\linewidth]{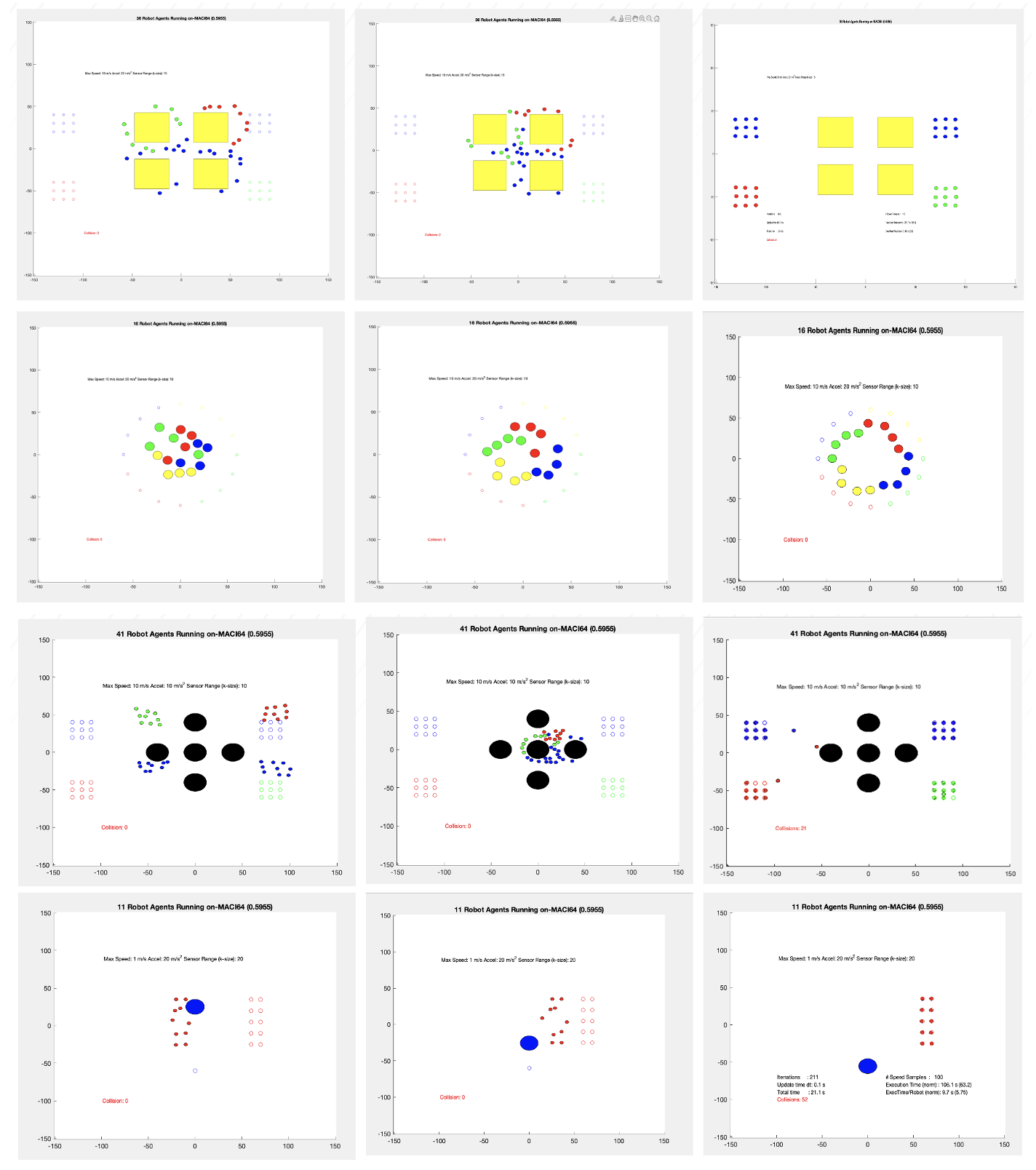}
	\caption{Simulated Experiments.  Top row: 4 groups of 9 robots navigating four square obstacles.  Second row: 4 Groups of 4 robots in a circle.  Third row: 4 groups of 9 robgots navigating five circular obstacles.  Fourth row:  10 robots in 5 rows of 2 robots each navigating a single dynamic circular obstacle.} 
	\label{fig:ex2} 
	\end{center}  
\end{figure}

\begin{figure}[h]
    \centering
	\begin{center} 
	\includegraphics[width=0.9\linewidth]{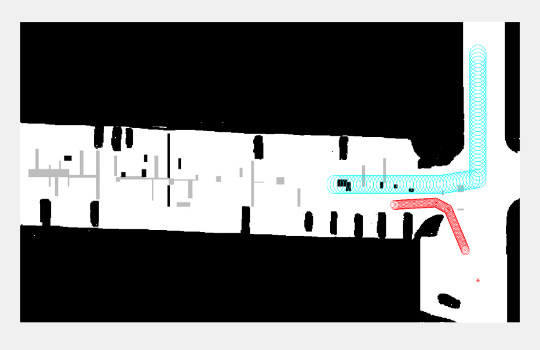}
	\caption{Mapped Environment for RVO} 
	\label{fig:rvo2} 
	\end{center}  
\end{figure}

\begin{figure}[h]
    \centering
	\begin{center} 
	\includegraphics[width=1\linewidth]{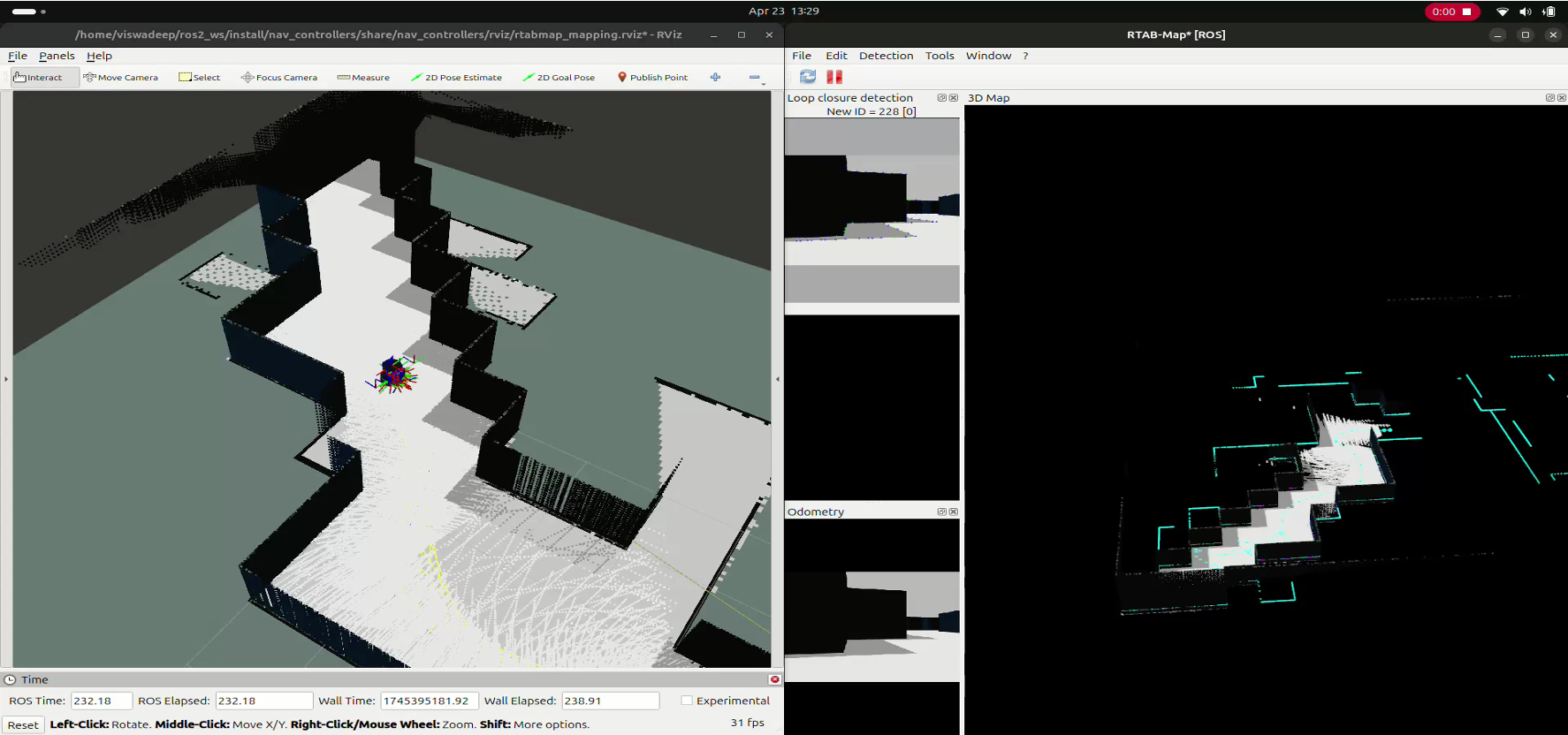}
	\caption{3D Maze Simulation in Gazebo} 
	\label{fig:maze} 
	\end{center}  
\end{figure}

\begin{figure}[h]
    \centering
	\begin{center} 
	\includegraphics[width=1\linewidth]{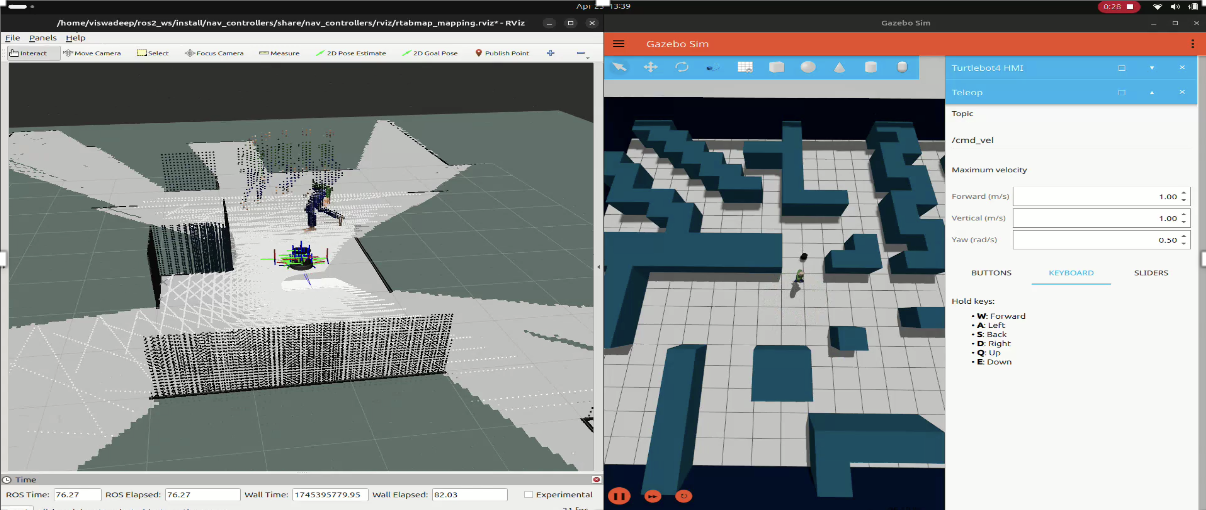}
	\caption{3D Simulation in Gazebo with Moving Person} 
	\label{fig:maze2} 
	\end{center}  
\end{figure}

\ \ \ To test the ORCA-FL algorithm, several simulated experiments were performed to assess how the ORCA-FLC algorithm performs with multi-agent and/or multi-obstacles.  All experiments involved multi-agents going from start positions to specified goals (circles with a perimeter the same color as the agent) in dynamic environments.  The first experiment involves 4 groups of 9 robots moving around four square obstacles to get to their goal points (Fig. \ref{fig:ex2} top row).  The second experiment involves placing 4 groups of N robots placed in a circle to move to their goal points (Fig. \ref{fig:ex2} second row).  The third experiments involves 4 groups of 9 robots moving around four circular obstacles (Fig. \ref{fig:ex2} third row)  The fourth experiment involves robots in a row moving to their goal by avoiding collision with a moving circular obstacle (Fig. \ref{fig:ex2} fourth row).  The fifth experiment involves robots in a row moving to their goal by avoiding collision with a square obstacle (not shown).  The difference in use between circles and squares is intended to capture how agents handle collision avoidance with non-polygons and polygons.
Simulations in a mapped environment were then performed.  Figure \ref{fig:rvo2} shows a simulation of the RVO algorithm.  With RVO, collision avoidance is only assured for 2 agents.   
Using the same mapped environment, ORCA-FL was simulated in a multi-agent environment where the start and goal points for each agent could be selected. 
\begin{figure}[h]
    \centering
	\begin{center} 
	\includegraphics[width=0.9\linewidth]{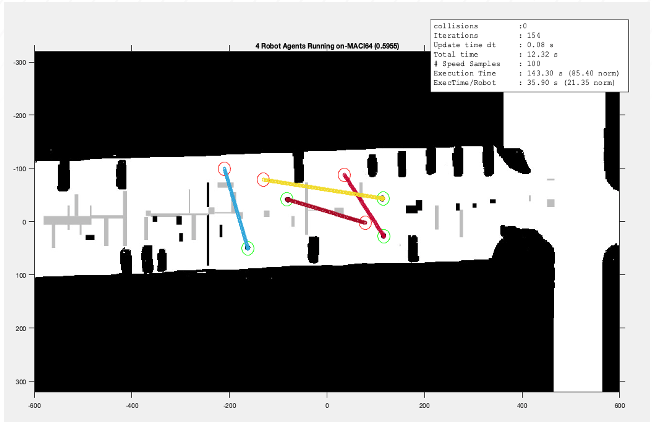}
	\caption{Mapped Environment for ORCA-FL} 
	\label{fig:flc2} 
	\end{center}  
\end{figure}   
    \ \ \ Using NAV2, a 3D maze environment was created in Gazebo/RVIZ.  A turtlebot with LIDAR, a depth camera, and odometry sensors (sensor fusion) was placed in the environment.  The red, blue, and green lines pointing outward from the turtlebot represent the various sensors.  On the right side of Figure \ref{fig:maze} is the point cloud representation of the maze captured by the LIDAR.   
    Figure \ref{fig:maze2} illustrates a simulation of the turtlebot in a maze with a walking/moving person and the sensing of that person using sensor fusion.

\section{Results}
    The experimental results are shown in Table \ref{tbl:results} for five experiments for ORCA and ORCA-FL using different maximum velocities of 10, 20 and 30 m/s.  We assume $dt=0.1$ and maximum acceleration $acc=20 \ m/s^{2}$.  Thus, the maximum speed differential $acc\cdot dt = 2 \ m/s^{2}$.
\begin{table}[h]
\scriptsize
\tiny
\caption{Experimental Results}
\label{tbl:results}
\begin{tabular}{ |c | c | c | c | c | }
 \hline
 \multicolumn{5}{|c|}{\textbf{ORCA-FL}  ($v_{max}$ = 10 m/s)} \\
 \hline
 Experiment & Num. Steps & Elapsed Time (s) & Num. Obs./Agents & Num. Collisions \\
 \hline
 1 (Robots in Circle)  & 196 &  19.6 &  0/16 & 0 \\
 2 (4 groups of 9: 4 square obs.) &  408 & 216.1 & 4/36 & 104  \\
 3 (Mov. square obstacle) & 220 &  70.8 & 1/10 & 20 \\
 4 (Mov. circular obstacle) & 195 & 42.1 & 1/10 & 43  \\
 5 (4 groups : 4 circ. obs.) : & 479 & 549.6 &  4/36  & 21  \\
 \hline
  \multicolumn{5}{|c|}{\textbf{ORCA} ($v_{max}$ = 10 m/s)} \\
 \hline
 Experiment & Num. Steps & Elapsed Time (s) & Num. bs./Agents & Num. Collisions \\
 \hline
 1 (Robots in Circle)  & 190 &  19 &  0/16 & 0 \\
 2 (4 groups of 9: square obs.) & 313  & 161.9  & 4/36 & 98    \\
 3 (Mov. square obstacle) & 109 & 24.7  & 1/10 & 0  \\
 4 (Mov. circular obstacle) & 158 & 28.6  & 1/10 &  21 \\
  5 (4 groups of 9: 4 circ. obs.) : & 440  & 257.3  &  4/36 & 120    \\  
 \hline 
  \multicolumn{5}{|c|}{\textbf{ORCA-FL} ($v_{max}$ = 20 m/s, $acc_{max} = 20$, sensor range (k size) = 15} \\
  \hline
 1 (Robots in Circle)  & 110  &  24.9 & 0/16 & 0   \\
 2 (4 groups of 9: 4 square obs.) & 314   & 208.7  & 4/36 & 25   \\
 3 (Mov. square obstacle) & 61 & 6.8  & 1/10 &  5 \\
 4 (Mov. circular obstacle) & 76  & 12.2  & 1/10 & 5   \\
 5 (4 groups : 4 circ. obs.) : & 225  & 72.5  &  4/36  & 80   \\
 \hline
  \multicolumn{5}{|c|}{\textbf{ORCA} ($v_{max}$ = 20 m/s} \\
 \hline
 Experiment & Num. Steps & Elapsed Time (s) & Num. bs./Agents & Num. Collisions \\
 \hline
 1 (Robots in Circle)  &  107 &  18.9  &  0/16 & 0  \\
 2 (4 groups of 9: 4 square obs.) & 286 & 113.5  & 4/36 & 18    \\
 3 (Mov. square obstacle) & 58 & 7.3  & 1/10 & 19  \\
 4 (Mov. circular obstacle) & 139  & 11.4  & 1/10 & 53  \\ 
  5 (4 groups of 9: 4 circ. obs.) : &  265  & 107.2  &  4/36 & 79    \\
 \hline 
  \multicolumn{5}{|c|}{\textbf{ORCA-FL} ($v_{max}$ = 30 m/s} \\
  \hline
 1 (Robots in Circle)  & 55  &  23.6 & 0/16 & 29   \\
 2 (4 groups of 9: 4 square obs.) &  -   & -   & 4/36 & 41 \\
 3 (Mov. square obstacle) &  58 & 8.8  & 1/10 &  3 \\
 4 (Mov. circular obstacle) & 49  & 7.3  & 1/10 & 3   \\
 5 (4 groups : 4 circ. obs.) : & - & -  &  4/36  & 73   \\
 \hline
  \multicolumn{5}{|c|}{\textbf{ORCA} ($v_{max}$ = 20 m/s} \\
 \hline
 Experiment & Num. Steps & Elapsed Time (s) & Num. bs./Agents & Num. Collisions \\
 \hline
 1 (Robots in Circle)  & 64  &  16.6 &  0/16 & 36  \\
 2 (4 groups of 9: 4 square obs.) & - &  - & 4/36 & 47 \\
 3 (Mov. square obstacle) & 58  & 6.7  & 1/10 & 3  \\
 4 (Mov. circular obstacle) &  75 &  8.5  & 1/10 & 3  \\ 
  5 (4 groups of 9: 4 circ. obs.) : & -   &  -  &  4/36 & 101    \\
\hline 
\end{tabular}
\end{table}
 The results indicate that at $v_{max}$ = 10 m/s, the standard ORCA outperforms ORCA-FL for all experiments except the 4 groups of 9 robots that need to move around 4 circular objects in which ORCA-FL only has 21 collisions while ORCA has 120 collisions.   At $v_{max} = 20 m/s$, ORCA-FL outperforms ORCA in the experiments of 10 robots in a row that need to avoid a single large moving square and circular obstacle, respectively. ORCA and ORCA-FL have no collisions at $v_{max} = 10 m/s$ and $v_{max} = 20 m/s$ for the 4 groups of 4 robots in a circle experiment.  However, at $v_{max} = 30 m/s$, ORCA-FL outperforms ORCA for the robot in a circle experiment (29 v. 36 collisions), in the 4 groups of 9 robots that need to move around four square obstacles experiment (41 v. 47 collisions), and in the 4 groups of 9 robots that need to move around four circular obstacle experiment (73 v. 101 collisions).  ORCA-FL and ORCA tie and have the same number of collisions, 3, for the 11 robots in a row experiment that need to avoid one large moving large square and circular obstacle, respectively. 
    
This suggests that the FLC provides more accuracy as the variability increases in the environment through velocity increases of all the other multi-agents and/or obstacles.  As velocity and acceleration decrease, ORCA performs better than ORCA-FL because there is less uncertainty and variation in the environment.  For example, at very slow or slow speeds, that is, 10 m/s or less, the usefulness of the collision avoidance parameter $\boldsymbol{u}$ and the expected velocity of the nearest obstacle / agent decreases, since there is more certainty that the collision avoidance responsibility should be shared equally and that the robot does not need to anticipate substantial increases in the expected velocity of the obstacles/agents in its neighborhood.  

In certain experiments at high velocities, the elapsed time and number of steps could not be computed and were not available (denotes as - in Table III) because the agents oscillated back and forth once it reached the goal point rather than stop and reach a steady state.  This may be due to the fact that at higher velocities, there could be noise introduced with feedback loops into the FLCs.  This requires further investigation and analysis.  The FLCs were used to optimize parameters used in the calculated trajectory of the obstacle over a short time horizon of $\Delta t$.  Thus, FLCS make decisions for autonomous local path planning.   The inputs and outputs to the FLCs depend on the robot type, sensed data, and environment.  For instance, if instead of a turtlebot, a car-like robot with four wheels or Ackermann steering was used, it would be appropriate to use FLCs to control steering direction and braking using velocity and steering angle inputs. 

\section{Fuzzy Optimization and Tuning}
    To improve the accuracy and efficiency of a FLC's response, one can optimize its performance through fine-tuning by adjusting parameters like membership functions and fuzzy rules to better suit the specific process being controlled.  Fine tuning can minimize discrepancies between the desired output and the actual output of the controller.  Particle swarm optimization, genetic algorithms, and simulated annealing are methods that can be used for fining tuning FLCs.  Various researchers have used fuzzy reinforcement learning and optimized fully logic controllers for obstacle avoidance and path planning for autonomous mobile robots \cite{Riman:2024, Ghaderi:2012, Lakhmissi:2019, Anam:2009, Deng:2004, Guan:2023}.  Allawi and Abdalla \cite{Allawi:2014} used an optimized Interval Type-2 fuzzy logic controller (IT2 FLS) using the artificial bee colony optimization algorithm (ABC) and the HRVO algorithm to handle collision avoidance.  
    
    To optimize the inputs and outputs of the FLCs, the genetic algorithm (GA) learning method can be used by specifying a population size (i.e. 200) and a maximum generation size (i.e. 25) in Matlab using the \textit{getTunableSettings} and \textit{tunefisOptions} functions.  The input data can be generated through Monte Carlo simulation of the inputs by assuming they follow an Ornstein-Uhleneck stochastic diffusion processes using $N$ simulations (i.e., 100,000) and a time step of $\Delta t = 1/N$.  The estimated mean velocities, distances, and accelerations can be used for their drift terms and their estimated volatility for the stochastic term driven by a Weiner process (Brownian motion).  This simulated input can be used in the FLCs to generate the rule outputs for optimization.  However, running the GA tuning is computationally expensive and depending on the number of simulations and GA parameters, can significant execution time.

    Introducing reinforcement learning (RL) into the algorithm to optimize the FLC may also improve the results by reducing the number of collisions at any given velocity.  In RL, an agent learns to optimize interactions in a dynamic environment through trial and error using a state-action feedback system where the agent receives a scalar value (reward) $R_{t}$ for each action taken at time $t$.  The goal of the agent is to learn a strategy for choosing an action $a_{t}$ after perceiving the state $s_{t}$ of the environment that maximizes the expected sum of discounted rewards.  The environment responds by giving the agent a scalar reinforcement signal, $R(s_{t})$ and transitioning into state $s_{t+1}$.  Thus, the agent follows a policy $\pi: (s,a) \rightarrow [-1,1]$, maximizing the action value Q function, a temporal difference reinforcement algorithm.  The update rule is
    \begin{multline}
        Q^{\pi}_{t+1}(s_{t},a_{t}) \leftarrow Q^{\pi}_{t}(s_{t},a_{t}) + \eta(R_{t+1} + \\ \gamma(\underset{a_{t}}{\text{max}}\ Q^{\pi}(s_{t+1},a_{t}) - Q^{\pi}(s_{t},a_{t})) 
        \label{eq:16}
    \end{multline}
    where $\eta$ is the learning rate and $\gamma$ is the discount factor.   An optimal policy is 
    \begin{equation}
        \pi^{*}(s) = \underset{a}{\text{arg max}} \ Q^{\pi}(s,a), \ \forall \ s \in \mathbb{S} 
    \end{equation}
    Thus, the optimal policy $\pi^{*} $consists of a series of actions that satisfy:
    \begin{equation}
        Q^{\pi^*}(s_{t},a_{t}) = \underset{\pi}{\text{max}}\ \mathbb{E} \bigg (\sum^{\infty}_{k=t}\gamma^{k-t}R_{k} \bigg )
    \end{equation}
    Q-learning \cite{Watkins:1992} can be used to tune the FIS. All state-action pairs are stored on a lookup table.  Action that move the agent closer to the goal receive a positive reward while rewards for other state-action pairs are negative valued.  For instance, 
    \begin{equation}
        R_{t+1} = \begin{cases}
                         +1 \ \ \ \text{if} \ s_{t+1} \text{is a goal state} \\
                         -1 \ \ \ \text{otherwise}
                   \end{cases}
        \label{eq:reinforce}
    \end{equation}
    The advantage of Q-function is that the agent can perform one-step lookahead search without knowing the one-step ahead reward and dynamic functions.  However, Q-learning is subject to the curse of dimensionality because the Q-function increases as the state domain increases.
    With fuzzy Q-learning, a Q value for multiple conclusions or consequents is associated with each rule.  Several competing conclusions are associated to each rule, and a quality value is assigned to each conclusion. The conclusion with the high quality (Q value) is used by the FIS to generate actions.  
      
    The fuzzy Q-learning (FQL) algorithm allows the training process to generate the best rules that maximize future reinforcements 
    \cite{Jouffle:1998}.  The original proposed rule-base using a zero order Takagi-Sugeno model has $N$ conclusions of the following form:
    \begin{align}
        \centering 
        \text{if} \ x\ \text{is} \ S_{i}, \text{then} \ y &= \text{action} \ a[i,1] \ \text{with quality} \ q[i,1] \nonumber  \\  
        \text{or} \  y &= \text{action} \ a[i,2] \ \text{with quality} \ q[i,2]  \nonumber  \\   
                        \vdots \nonumber  \\   
        \text{or} \ y &= \text{action} \ a[i,N] \ \text{with quality} \ q[i,N]   \nonumber 
    \end{align} 
    where $q[i,j], i=1,\dots,M, \ j=1,\dots,N$, are potential solutions with values initialized to 0 at the beginning of training.  During the learning process, the conclusion part of each rule is selected by means of an exploration-exploitation policy (EEP), where $EEP(i) \in \{1,\dots,N\}$ such as $\epsilon$-greedy action selection \cite{Ghaderi:2012}: \\ \\
    $A^{\epsilon-greedy}_{t} = 
    \begin{cases}
    \underset{a \in A(s_{t})}{\text{arg max}} \ Q(s_{t},a_{t} \ \text{with probability} \ \epsilon \\
    \text{random action} \ a(s_{t}) \ \text{with probability} \ 1-\epsilon \\
    \end{cases} \\ \\
    $
    The inferred action is given as:
    \begin{equation}
        A(s) = \sum_{i=1}^{m}w_{i}(s)a[i,EEP(i)]
        \label{eq:20}
    \end{equation}
    and the quality (Q-value) of the inferred action is:
    \begin{equation}
        \hat{Q}(s,A(s)) = \sum_{i=1}^{m}w_{i}(s)q[i,EEP(i)]
        \label{eq:21}
    \end{equation}
    where 
    $w_{i} = \alpha_{i}(s)/\sum^{M}_{i=1} \alpha_{i}(s)$ and    
    $\alpha_{i}(s)$ is the (truth) value of the $i$th rule.  Let $q[i,i^{*}] = \underset{j \leq m}{\text{max}} \ q[i,j]$.
    The optimal value of state $s$ is then \cite{Anam:2009}:
    \begin{equation}
    Q^{*}(s,a) = \frac{\sum^{N}_{i=1}\alpha_{i}\cdot q(i,i^{*})}{\sum^{N}_{i=1}\alpha_{i}(s)}
    \end{equation}  
    If $s_{t+1}$ is the new state and $E_{t+1}$ is the reinforcement signal using equation \ref{eq:reinforce}, then $Q(s,a)$ can be updated using equations \ref{eq:16}, \ref{eq:20}, and \ref{eq:21}.  The difference between the old and new Q-values can be viewed as an error signal: 
    $\Delta Q = Q^{\pi}_{t+1}(s_{t},a_{t}) - Q^{\pi}_{t}(s_{t},a_{t}) = R_{t+1} + \gamma Q^{*}(s_{t+1},a_{t}) - Q^{\pi}(s_{t},a_{t})$ that can be used to update the action q-values.  Using ordinary gradient descent, we get:
    \begin{equation}
      \Delta q[i,i_{a}] = \eta \Delta Q^{\pi} \frac{\alpha_{i}(s)}{\sum^{N}_{i=1}\alpha_{i}(s)} 
      \label{eq:update}
    \end{equation}
    After a training time period, the robot should choose for each fuzzy-rule, the best conclusion corresponding to the best Q-function $q[i,j] = q[i,i_{a}]$.
    Learning can be improved and accelerated by combining Q-learning and the temporal-difference (TD($\lambda)$) method \cite{Glorennec:1997, Tanner:2005, Sutton:1988}, generating eligibility traces (new memories) $e[i,j]$ of an action y: \\ 
    \begin{equation}
    e[i,j] \leftarrow 
    \begin{cases}
        \lambda \gamma e[i,j] + w_{i} \ \text{if} \ j = i_{a} \\
        \lambda \gamma e[i,j] \ \ \ \ \ \ \ \ \text{otherwise}
    \end{cases}
    \end{equation}
    where e[i,j] is initialized to 0.  Therefore, updating equation \ref{eq:update} yields:
    \begin{equation}
            \Delta q[i,j] = \eta \Delta Q^{\pi} e[i,j]
    \end{equation}
    More formally, in the semi-gradient TD($\lambda$) algorithm, a new memory vector known as an eligibility trace $\textbf{z}_{t} \in \mathcal{R}^{d}$, $\boldsymbol{z}_{-1} = 0$, follows the following process for estimating $\hat{v} \approx Q^{\pi}$. 
    \begin{algorithm}
        \caption{\small \textbf{Semi-gradient TD($\lambda$)}}
    	\label{alg:td}
	   \scriptsize
	    \algsetup{linenosize=\small}
        Input: the policy $\pi$ to be evaluated and $\hat{v}$ \\
        Initialize value-function weights \textbf{w} = \textbf{0} \\
        \While{$\text{terminal state (goal) is not reached}$} { 
        \text{Choose} \ $A \sim \pi(\cdot \lvert S)$ \\
        \text{Take action $A$, observe $R$, s'} \\
        $\boldsymbol{z}_{t} = \lambda \gamma \boldsymbol{z}_{t-1} + \nabla \hat{v}(s_{t},\boldsymbol{w}_{t})$ \nonumber \\
        $\delta_{t} = R_{t+1} + \gamma\hat{v}(s_{t+1},\textbf{w}_{t}) - \hat{v}(s_{t},\textbf{w}_{t})$ \nonumber \\  
        $\boldsymbol{w}_{t+1} = \boldsymbol{w}_{t} + \alpha \delta_{t}\boldsymbol{z}_{t}$ \nonumber \nonumber \\
        $s \leftarrow s'$ \\ 
        }
        \textbf{end}
    \end{algorithm} \\
    $\boldsymbol{w} \in \mathcal{R}^{d}$ is a weight vector and $\hat{v}$ is a differentiable function where $\hat{v} : S \times R^{d} \rightarrow \mathbb{R}$ such that $\hat{v}$ ($s_{terminal},\cdot$) = 0.
    For each time step, the algorithm decays each component by $\lambda \gamma$ and the trace for the current state is incremented by 1, leading to an \textit{accumulating} trace \cite{Sutton:1989}.  
 
    Both GA and FQL are based on stochastic search: mutation and cross-over for GA And exploration for FQL.  However, with GA optimization, the fitness function is evaluated only once at the end of a trial.  In contrast, with FQL, the Q-values are evaluated online.  The performance for the entire set of conclusions is known with GA whereas FQL examines the quality of individual actions and thus provides more acute insight into learning and acquired knowledge.  Deep reinforcement learning and Q networks have also been proposed for mobile robot obstacle avoidance as well as Kohonen neural networks \cite{Feng:2018, Macek:2002, Li:2024}.  These networks adapt well to large state spaces, high-dimensional input spaces, and learn optimal strategies through interaction with the environment. However, fuzzy Q-learning is adept at handling complex, uncertain, and noisy dynamic environments. 

    The fuzzy reinforcement learning (FQRL) algorithm is given in Algorithm 3.
    FQL can be incorporated into ORCA-FL algorithm by using Q tables with FLC1 and FLC2.  The state variables for FLC1 are distance, velocity, and acceleration.  Depending on the states, the shared avoidance collision output $\textbf{u}$ will have different conclusions leading to different actions with corresponding q-values.  Likewise, depending on the inputs (states) for FLC2, the predicted/expected velocity of the obstacle will have different conclusions leading to different actions with corresponding q-values.  The robot will choose the actions with the highest q-values based on the $\epsilon-greedy$ action selection method and the Q-tables will be updated according to line 12 of the algorithm. 
                
\section{Conclusion}
    The proposed ORCA-FL is a hybrid algorithm combining a local path planner with a heuristic method.   ORCA-FL improves on RVO because it can be used in a multi-agent environment and does not result in deadlocks and/or oscillations.  Fuzzy logic controllers  can improve obstacle avoidance by deciding the optimal collision avoidance responsibility parameter and the expected/predicted velocity factor of the nearest obstacle.  FLCs use linguistic rules to handle uncertainty and imprecision in the sensed measured values in the environment.  Contrary to the results in \cite{Romlay:2023}, the results did not strictly outperform ORCA except when the velocity exceeds a certain threshold.   The threshold can be determined by using stochastic gradient descent to find the velocity that leads to a global minimum (of collisions) on the FIS surface using fuzzy inputs.

\begin{algorithm}
	\label{alg:rrtstaralg}
	\scriptsize
	\algsetup{linenosize=\small}
	\caption{\small \textbf{FQL Algorithm}} 
    $\textbf{Input} \ \eta : \gamma$ \\ 
    $\text{Initialize:} Q(s,a) \text{table with zeros}, R_{tot} \text{with 0}$ \\
    \For{$i \in 1,\dots,N$}  { 
        $\text{Initialize} \ s$ \\
        \For{$j \in 1,\dots,M$} { 
            $\text{Choose action} \ a_{j} \ \text{at state} \ s \ \text{using policy derived from} \ Q$ \\
            $\text{using} \ \epsilon-greedy \ \text{action selection method}$ \\
            $\text{Take action} \ a_{j} \ \text{and observe} \ R_{j}, s_{j+1} \ \text{from environment}$
        }
        $\textbf{end}$ \\
        \If{$\sum^{M}_{k=1}R_{k} > R_{tot}$} { 
            \For{$j \in 1,\dots,M$} { 
                $Q(s_{j},a_{j}) \leftarrow Q(s_{j},a_{j}) + \eta[R_{j} + \gamma \underset{a}{\text{max}} \ Q(s_{j+1},a_{j}) - Q(s_{j},a_{j})]$ \\
            }
            $\textbf{end}$ \\
            $\text{Initialize} \ s$ \\
             \For{$j \in 1,\dots,M$} {  
                $\text{Choose action} \ a_{j} \ \text{at state} \ s \ \text{using policy derived from} \ Q \ (\epsilon-greedy \ \text{action selection method})$ \\
                $\text{Take action} \ a_{j}, \ \text{observe} R_{j},s_{j+1} \ \text{from environment}$
            }
            $\textbf{end}$ \\
            $R_{tot} = \sum^{M}_{k=1}R_{k}$ \\
        }
        $\textbf{end}$ \\
    } 
    $\textbf{end}$ \\
\end{algorithm}

\section{Future Research}
    Future work will require testing the performance of ORCA-FL in different map environments and using different fuzzy logic membership functions.  The algorithm needs to be compared to other obstacle avoidance methods including dynamic window approach (DWA) \cite{Fox:1997, Coissac:2023} and timed elastic band (TEB). In addition Field experiments need to be performed in actual multi-agent maze/map environments replicating the simulated maps. Finally, fuzzy logic tuning and optimization using FQL and GA needs to be implemented in ORCA-FL and tested to evaluate and compare the new algorithm (ORCA-FQL) collision performance with ORCA-FL, ORCA, and other obstacle avoidance algorithms.  
    
\bibliographystyle{IEEEtran}
\bibliography{bibl.bib}

\vspace{12pt}

\end{document}